# Software Agents with Concerns of Their Own

Luís Botelho        Luís Nunes        Ricardo Ribeiro        Rui J. Lopes

## Author Notes

Luís Botelho, Luís Nunes and Rui J. Lopes, Instituto de Telecomunicações (IT-IUL) and Department of Information Science and Technology of Instituto Universitário de Lisboa (ISCTE-IUL), Lisbon, Portugal

Ricardo Ribeiro, Instituto de Engenharia de Sistemas e Computadores (INESC-ID) and Department of Information Science and Technology of Instituto Universitário de Lisboa (ISCTE-IUL), Lisbon, Portugal

Correspondence concerning this article should be addressed to Luís Botelho, of the Department of Information Science and Technology of Instituto Universitário de Lisboa (ISCTE-IUL), Av. das Forças Armadas 1649-026, Lisboa, Portugal. Phone: +351 963 405 048

Contact: Luis.Botelho@iscte-iul.pt

# ABSTRACT

We claim that it is possible to have artificial software agents for which their actions and the world they inhabit have *first-person* or intrinsic meanings. The *first-person* or intrinsic meaning of an entity to a system is defined as its relation with the system's goals and capabilities, given the properties of the environment in which it operates, as recognized by the system. Therefore, for a system to develop *first-person* meanings, it must see itself as a goal-directed actor, facing limitations and opportunities dictated by its own capabilities, and by the properties of the environment.

The first part of the article discusses this claim in the context of arguments against and proposals addressing the development of computer programs with *first-person* meanings. A set of definitions is also presented, most importantly the concepts of cold and phenomenal *first-person* meanings.

The second part of the paper presents preliminary proposals and achievements, resulting of actual software implementations, within a research approach that aims to develop software agents that intrinsically understand their actions and what happens to them. As a result, an agent with no *a priori* notion of its goals, capabilities, or the properties of its environment acquires all these notions by observing itself in action. The cold *first-person* meanings of the agent's actions and of what happens to it are defined using these acquired notions.

Although not solving the full problem of *first-person* meanings, the proposed approach and preliminary results allow us some confidence to address the problems yet to be considered, in particular the phenomenal aspect of *first-person* meanings.

# Software Agents with Concerns of their Own

## 1   Introduction

In a debate that came to be better known as the symbol grounding problem [1], several scientists have expressed their opinions that artificial intelligence would never lead to actual intelligent systems (e.g., [2, 3, 4, 5]). For these authors, artificial systems do not understand what happens to them, the symbols they possess, and what they do. That is, even when an artificial system exhibits a meaningful behaviour, that meaningfulness is in the head of the observer; it is not an intrinsic interpretation of the artificial system itself.

We argue that it is possible to develop computer programs for which, objects, events, and situations have intrinsic meanings to the computer program, that is, *first-person* meanings. We use italics because we are applying to machines, an expression explicitly intended for people.

Several of our arguments rely on the view that at least simple non-conscious living beings and artificial systems share the fundamental property of being purely rule-following mechanisms. That is, their behaviour in a given state (internal and external) is completely determined by the interaction between the natural or artificial forces applicable to their inner workings and the properties of that state. Relevant forces applicable to artificial systems may include the instructions of a computer program. Relevant forces applicable to living beings include, among others, electrical and gravitational forces. From this point of view, at least simple rule-following living beings face the same objections, regarding the possibility of developing *first-person* meanings, as those presented to artificial intelligence by their critics, especially that the meaningfulness of the behaviour of simple organisms is in the head of the observer. And, just as *first-person* meanings have emerged through evolution from non-conscious beings, we believe that they can also develop in artificial systems.

In spite of having presented the parallel between simple non-conscious rule following organisms and simple rule following artificial systems, we believe that the behavior of even complex adaptive organisms such as human beings emerges of rule following mechanisms. For instance, Badcock, Friston and Ramstead [6] provide a completely mechanist causal account of the human brain, at the level of information theory. For them, the human brain is an adaptive system that actively minimizes the free energy (and indirectly, the entropy) of human sensory and physical states by producing self-fulfilling action-perception cycles that emerge from dynamic interactions between hierarchically organized neurocognitive mechanisms. This is a pure rule following model, originally inspired in thermodynamics, linking brain, mind and behavior in human beings. Wiggins [7] also presents a similar purely mechanist account of the human cognition, according to which, mind processes information driven by the information theoretical principal of information efficiency.

Therefore, if persons have or develop a first-person perspective, in spite the fact that their mind and behavior emerge of rule following mechanisms, there is nothing in principle that precludes the same from happening with computer programs. As noted by Wagner [8], nothing intrinsically restricts personhood to human beings. Conceptually, personhood allows for the possibility of non-human persons that are not members of our species.

This article presents our arguments and claims, our proposed research framework, and its preliminary results. It describes a set of conceptually defined, computationally implemented, and demonstrated concepts and mechanisms that support the development of *first-person* meanings (in a particular sense to which we call cold *first-person* meanings), and the enactment of meaningful behaviour. The described mechanisms and achieved results are presented within the context of the described research framework.

The next section presents some terminology, emphasizes the main contribution of the article, and discusses the five assumptions of our proposed research framework. That research framework and the defined, implemented, and demonstrated concepts, mechanisms, and results are detailed in section 3.

## 2 Research Assumptions, Terminology, Contribution, and Discussion

This article discusses several concepts whose definitions are not widely accepted. In some cases, each author uses his/her own definitions. In other cases, not even a definition is presented in the hope that the common sense understanding of the concept is enough for the purpose of the discussion. In the next subsection, we explain the sense in which several concepts are used in the article, and we emphasize the main contribution of the article. Then, we describe a brief example that illustrates the way we use those concepts. Finally, we present and discuss our research assumptions.

The discussion and analysis present in this section are put forward in an informal fashion. Our approach is rigorously formalized in section 3.

## 2.1 Background Concepts and Main Innovative Contribution

This section starts with a few clarifications regarding certain concepts and the way they are used in this article. First, we discuss the concept of goals. After that, we present several distinctions about consciousness and related concepts, especially *first-person* meanings.

<u>Goals and other motives</u>. One of the major claims of this article is that to develop (or to have) *first-person* meanings, a system needs to see itself as having goals or other motives. Given that the main focus of the article is not the distinction among different kinds of motives (e.g., value systems, desires, goals and intentions), for the sake of simplicity, we will use the term goals.

Additionally, systems may see themselves as if they had goals (or other motives), even if they do not actually have goals. As explained by Daniel Dennett [9], the way a system works can be described as if it had mental properties (e.g., goals and beliefs) in spite the fact that its behaviour may be completely determined by applicable forces at the physical level. This mental level of description – the intentional stance – might be appropriate for understanding some of the higher-level properties of the system. Thus, although we may describe a system as having goals, it doesn't necessarily mean that the system's behaviour is actually determined by goals. Therefore, the remainder of the article will always refer to the system's goals whether those goals really determine the system's behaviour or they are just a way the system understands itself.

The meaning of goals for a system depends on its architecture. Some architectures are not capable of seeing themselves as having goals, although an external observer may describe their behaviour as goal-directed. For systems with such an architecture, goals are totally meaningless.

Some other architectures allow the system to understand itself and the world as if it had goals, but otherwise do not allow them to shape the other aspects of its behaviour because of its goals. In these cases, goals provide a way for the system to find meaning, to understand what happens, to know what is and what is not important. Aside from that, goals do not have other impacts in the system's behaviour.

Finally, there are other more sophisticated architectures that allow a system both to see itself and the world as if it had goals and also to shape its behaviour because of its goals. In this paper, we say the system adopts its goals. In the line of the work of Anderson and Perlis [10], for these systems, having goals means to be capable of consistently conditioning their behaviour to directly or indirectly achieving their goals.

The present article is mainly focused on the process that may be used by a rule following system to see itself as goal-directed and, after that, developing *first-person* meanings. In addition to this main objective, the paper describes and demonstrates a possible architecture capable of changing its behaviour after the system sees itself as goal directed.

Without any intention of providing a comprehensive view of consciousness, we need to present a few distinctions that will be used to explain other concepts. Consciousness has several manifestations and properties such as self-location and the experienced direction of the *first-person* perspective [11], global bodily ownership [12], the autobiographic sense of the self [13], awareness [14], reasoning [14], and language [15]. We will use the term "awareness" and derivatives to refer only to the knowledge or information one has by virtue of being conscious. In the sense that it is used hereunder, awareness is not meant to include for instance declarative reasoning and language. The special kind of awareness referring to the self would be termed self-awareness.

As other manifestations of consciousness, awareness has a phenomenal component: when we are aware, we have phenomenal experiences, we have sensations with certain qualities, we feel. The phenomenal component of consciousness was termed the hard problem of consciousness by David Chalmers [16]. In the scope of this article, it is helpful to distinguish the information, stripped of any sensations, to which a sentient individual has access when he or she is aware of something. For instance, when a parent is aware of the presence of his/her children, he/she experiences a sensation (phenomenal) and he/she knows that (he/she has access to the information that) they are present (cold).

In the remainder of this article, we will use the expressions *phenomenal awareness* to refer to the sensations one feels because one is aware of something, and *cold awareness* to refer to the information, stripped of sensations, to which a sentient individual has access because it is aware of something.

Finally, consciousness is also associated with a sense of continuity. When a person experiences and has access to information about the object of consciousness, for instance the image of an apple, the qualities of the experience and the information the person had access to, during the experience, will persistently be associated with the same object of consciousness (and to others, such as the members of the same class). This sense of continuity of consciousness is possibly ensured by autobiographic memory. Damásio [13] distinguishes core consciousness from extended consciousness. Each time a person encounters an object of which he or she becomes aware, there is a pulse of core consciousness. Each pulse of core consciousness is a short duration event; alone, it does not ensure the sense of continuity. Only extended consciousness, possibly relying on autobiographic or episodic memory, provides that sense of continuity.

*First-person* meanings, intrinsic meanings, and concerns. A system has (or develops) *first-person* meanings of things, situations, or events, if those things, situations, or events have (or acquire) meaning for the system itself. This expression is used in contrast with what happens when the meaning is in the mind of the observer (*third-person* meanings), but not in the system itself. In computer science, when a token has (or acquires) meaning, it is (or becomes) a symbol. We can think of the meaning of a symbol as the object, the situation, or the event that it refers to. In this article, we are interested on a different aspect of meaning, namely the importance of the object, situation, or event to the system itself. Thus, in this view, given that the importance of an object for a system reflects the relation of the object with the system's goals (or values, or any other source of motivation such as the need to survive), as recognized by the system, it is important to stress that, in the sense that we use the expression in this article, *first-person* meanings require the owners of those meanings to know that they have goals and recognize them.

Given the above characterization, the *first-person* meanings of objects, situations, or events reflect a value system the agent is aware of (recognizes). In this sense, when an object, an event, or a situation has *first-person* meanings for an agent, that agent has concerns about that object, event, or situation.

Cold *first-person* meanings, phenomenal *first-person* meanings, and consciousness. A system has (or develops) *first-person* meanings of things, situations, or events only if the system is aware of (i.e., recognizes) the intrinsic importance of those things, situations, and events to itself. If the recognition of such intrinsic importance results from mere cold awareness, we say the agent has cold *first-person* meanings or cold concerns. However, if the recognition of such intrinsic importance is the result of phenomenal awareness, if it is experienced, we say the system has phenomenal *first-person* meanings or phenomenal concerns. It is important to stress that we do not claim that the cold awareness involved in cold *first-person* meanings ensures the sense of continuity often associated with consciousness.

We claim that (cold / phenomenal) *first-person* meanings of objects, situations, or events are a manifestation of (cold / phenomenal) consciousness in the sense that we require that the system is aware of the importance of those objects, situations or events to itself. However, we do not equate *first-person* meanings with consciousness because consciousness has other manifestations such as autobiographic sense of the self, declarative reasoning, and language.

Given the distinctions presented so far, we are in the position of describing the main contribution of the article and also to clearly identify some aspects of the general problem that are not addressed in it. First of all, we restate that this article is concerned only with one aspect of *first-person* meanings, namely the importance of objects, situations, and events to the system. This is a difference regarding other work addressing *first-person* meanings, for example the work by Harnad [1], Cangelosi, Greco, and Harnad [17], Vogt [18], Machado and Botelho [19], Steels [20, 21], and Raue et al. [22], which focuses on the symbol reference problem, namely the problem of directly or indirectly grounding each symbol of an artificial system on the object, situation, or event referred to by the symbol.

We assume that goals, value systems, or other sources of motivation (such as the need to stay alive and the need to preserve our genes) are sources of importance. If an event facilitates or impairs the achievement of someone's goals, then that event is important for that someone. If a situation is unrelated to someone's goals, then the situation is unimportant to that someone. That is, *first-person* meanings or concerns require that the system sees itself as having goals.

Possibly the most significant innovative contributions of this article are *(i)* recognizing that, for the purpose of developing *first-person* meanings, the system must see itself as being goal-directed; and *(ii)* providing an approach for a program with no explicit notion of goals to discover that it might have goals and thereafter start interpreting its behaviour and what happens to it as if it were goal-directed. To the best of our knowledge, no other work has ever addressed the problem of *first-person* meanings in this way.

We claim the above contribution to be significant given that it represents a way around a plethora of objections about *first-personhood*, living beings, materialism, and computer programs. In fact, it has been argued (e.g., [23, 24]) that only living beings can develop intrinsic meanings because only living beings have intrinsic motives (the need to survive). To those arguments we reply that all that happens with simple organisms as bacteria is acting as determined by the interplay between their state (internal or external) and applicable natural forces – similarly to computer programs, they are just rule-following systems. However, this position faces a second sort of objections.

Searle [2], for example, says that (syntactic) rule-following systems cannot generate meanings (semantics) because semantics cannot arise from syntax. It is maybe this argument that reveals the potential importance of our contribution. The possibility of *first-person* meanings does not elude the rule-following nature of organisms and computer programs because the system does not have to exceed its nature, if it is to develop *first-person* meanings. In our view, it has to interpret itself as being goal-directed (even if its behaviour is only following rules). That is, *first-personhood* is within reach of rule-following systems.

Although required, seeing itself as goal-directed and recognizing its own goals is not enough to develop *first-person* meanings. If the system could always achieve all its assumed goals, it would not be concerned about them because there would not be true obstacles to their achievement. The importance of things, situations, and events is also dependent on the system's capabilities and on the properties of its environment. This article contributes to realize (or, at least, to emphasize) the described relationship and to provide the means for a computer program, originally with no notion of goals, capabilities, or the constraints and opportunities created by its environment, to learn them just by observing itself in action.

We claim that the presented contributions tackle the problem of developing cold *first-person* meanings, in the sense that it enables an agent to start interpreting its behaviour and what happens to it as if it were a goal-directed constrained entity. That is, the agent has cold awareness of the relation between objects, situations, and events and the possibility of achieving its goals, given its limited capabilities, and the constraints and opportunities created by the environment.

Although recognizing that solving the full problem of *first-person* meanings requires phenomenal awareness, the present work does not contribute to solve this more difficult problem. This will be an important object of future research.

The enumerated contributions are materialized as a set of concept definitions and algorithms that allow the agent to acquire cold *first-person* meanings.

The agent's sensors provide information regarding each state of the world and the actions it performs, leading from one state to the next. The information provided by the sensors is represented by predicates Proposition/1 and Action/1. Facts of the predicate Proposition/1, such as Proposition(On(A, B)) (i.e., On(A, B) is a proposition), describe each observed state. Facts of the predicate Action/1, such as Action(Move(A, 1, 2)), describe the performed actions. The agent also learns other properties of the entities of the domain, namely the concepts of static proposition, fluent proposition, desired proposition, and mandatory proposition.

The agent learns that its behaviour could be interpreted as if it were goal directed. Goal(*PropSet*) is the fact used by the agent to represent the goals it learns to have. The agent behaviour could be described as if the agent were striving to achieve a state in which the propositions in *PropSet* are true.

The agent learns the meaning of its actions. Precond(*Act*) represents the set of preconditions of action *Act*. PosEffects(*Act*) and NegEffects(*Act*) represent the sets of propositions that become true and false by virtue of executing action *Act*. ValidityCondition(*Act*) represents a logical condition that constraints the arguments that can be used in action *Act*, that is, it constrains the objects to which the action can be applied.

The agent also learns constraints that arise of the interplay between its environment, its goals, and its capabilities. MustPrecede(*Prop$_1$*, *Prop$_2$*) represents the fact that the proposition *Prop$_1$* must precede *Prop$_2$*, or else the agent will not be capable of achieving its goals.

Finally, and most importantly, the agent learns to explain its own actions in terms of its learned goals and in terms of the requirements of other actions it also performed. The predicates Achieved/3 and Contributed/3 represent what was achieved by the agent, when it performed a given action, and what was the contribution of an observed proposition to the actions executed by the agent. Achieved(*State*, *Act*, *PropsSet*) means that the action *Act*, executed in state *State* achieved all the relevant propositions in *PropSet*. These propositions are relevant for the agent because they are either goals or preconditions actually used by the agent to perform subsequent actions. Contributed(*State*, *Prop*, *Act*) means that proposition *Prop*, true in state *State*, actually contributed to action *Act*, which was executed in that *State* or in a future state.

Achieved/3 and Contributed/3 represent the *first-person* meanings of actions and propositions performed or observed by the agent in specific behaviours, by showing their contribution to the path leading to the achievement of its goals. The proposed approach does not address *first-person* meanings of objects; just of propositions and actions.

The agent learns all the mentioned concepts from scratch, that is, starting with nothing at all, only through observation of its action in the world. Before using the proposed algorithms, the agent does not know the propositions that describe the world or the ones that could describe it; it does not know anything about its actions, it does not know the goals it has or the ones it could have. The agent does also not know anything about precedence relations to which it must comply if it wants to achieve its goals. Finally, it does not know anything about the importance of specific observed facts for its action, or the reasons why it acted in a certain way. The agent learns all these, using the algorithms and deduction axioms described in subsection 3.1. The mentioned algorithms (although with some application restrictions) are absolutely independent of the application domain. The next section relates the learned concepts with an example.

## 2.2 Brief Example

We briefly present a short imagined example that illustrates our point of view and concepts.

A robot was programmed to weld the external panel of car doors and then to assemble the doors to the car. It has no goals because it is just following the instructions contained in its control program. It also does not have any notions of its capabilities nor of the properties of the environment. Using the proposed algorithms, the robot starts seeing itself as if it were motivated to achieve a situation in which it has assembled the doors to the car and has welded the external panels to the doors. We stress the fact that using the proposed algorithms leads the agent to discover its goals only in the cold sense.

In addition to believing that it has these goals, the agent also becomes aware of its capabilities, for instance, assembling a door to the car, grabbing the weld torch and using it to weld door panels, the conditions under which they can be used, and the effects they produce in the world (Precond/1, PosEffects/1, NegEffects/1 and ValidityCondition/1).

The robot also discovers properties of its task, such as that it must weld the panel to the door before it can assemble the door to the car (MustPrecede/2) because, although having the door panel welded is not a pre-condition to assemble the door, it cannot weld the door panel if the door is already assembled to the car.

In the eyes of the robot, a situation in which it has assembled the doors to the car and has welded the external panels to the doors means that it has fulfilled its goals – it is positively valued by the robot.

In a given behaviour (a specific sequence of actions), the welding robot picks up the welding torch, welds the external panel $P_1$ to the car door $D_1$, and assembles $D_1$ to the car. In this specific behaviour, the executed action of picking up the welding torch is meaningful to the robot because it achieved (Achieved/3) a proposition (holding the welding torch) that, although not one of its goals, contributed (Contributed/3) to executing the subsequent action of welding $P_1$ to $D_1$. The action of welding $P_1$ to $D_1$ is also meaningful to the robot because it achieved (Achieved/3) one of its goals (having welded $P_1$ to $D_1$). Having achieved a state in which $P_1$ is welded to $D_1$ before the state in which $D_1$ is assembled to the car is also meaningful to the robot because it satisfies the precedence relation (MustPrecede/2) according to which having the external panel welded to the door must occur before having the door assembled to the car.

The door, the door panel, the car, the welding station are meaningful, but our proposal does not address the meaning of objects; just the meanings of propositions and actions.

A situation in which the welding station is broken is also meaningful to the robot because it knows that it will not be capable of achieving its goals (in work developed after the one described in this article, we proposed algorithms that determine when a proposition prevents another one from occurring). This situation is negatively valued by the robot.

If the robot does not have feelings, in the phenomenal sense, the described meanings (positive and negative) are cold *first-person* meanings – the robot has cold concerns about those meaningful situations and events.

If the robot phenomenally feels the situation in which the welding station is broken, the situation has phenomenal *first-person* meaning to the robot – the robot has phenomenal concerns about it.

Finally, the car engine, the car paint, the situation in which the engine is not assembled to the car, or the event in which another robot damaged the car engine, for example, have no importance to the robot in the light of its learned goals. The robot does not have any concerns about the car engine or the other robots.

Section 3 describes the domain-independent definitions and algorithms used by an agent with no prior notion of goals, capabilities, or the properties of the domain, to become aware of all this information. After that, the agent is capable of explaining the meaning of events and situations, which, from its own point of view, acquire importance with respect to its assumed goals and capabilities, and the properties of the domain. Results are presented for a demonstration scenario.

## 2.3  Research Assumptions

Without a test, it is impossible to rigorously determine if a specific computer program has developed *first-person* meanings. Unfortunately, such a test has not yet been developed. Two major difficulties are hindering its development. The first difficulty arises from our insufficient understanding of what it means to have *first-person* meanings.

 The second difficulty comes from the fact that, even if it were possible to define a test for *first-person* meanings applicable to animals, it could happen that the same test would not be applicable to computer programs. To see that this may be the case, consider the mirror test for self-awareness [25]. This test was created by psychologist Gordon Gallup Jr. to determine whether a non-human animal possesses the ability of self-recognition, and it has been used as a self-awareness test. If an animal, in the presence of a mirror, is capable of adjusting its behaviour towards itself (e.g., touching a marked part of its body that becomes visible only in the mirror), then the animal recognizes itself in the mirror, which, for some, is evidence that the animal is self-aware.

If a computer program passes an adapted version of this test, it would not necessarily mean that the program is self-aware because it could have been programmed to exhibit the desired kind of behaviour without recognizing itself in the mirror. Possibly, this test would reveal that a program is self-aware only if we knew the program was defined in a certain way. Gold and Scassellati [26] developed an unsupervised learning algorithm that enables a robot to reliably distinguish its own moving parts from those of others, just by looking at a mirror. Although the authors explicitly state they do not claim that their robot is conscious, we may be willing to interpret this experiment as revealing self-awareness. However, this would happen only because we know that the robot uses an unsupervised learning algorithm – it was not explicitly programmed to act as if it recognized itself.  We still need better tests that do not rely on our knowledge of the way the program is written.

Without a test directed at determining the existence of *first-person* meanings, the discussion about the possibility of having computer programs with *first-person* meanings and the discussion about the way to achieve them cannot be more than just argumentative. We are aware of the arguments used by detractors of artificial intelligence. However, those arguments, as well as ours, may have flaws. Nevertheless, we think that pursuing the research objective of having computer programs with *first-person* meanings, even though it might be unachievable (which we do not believe), will have a beneficial impact on both our understanding of the problem and the possibility to create more effective computer programs capable of facing more complex problems.

We firmly believe that it is possible to have computer programs with *first-person* meanings. Our point of view and the research we have been doing is based on the following research assumptions:

1. It is possible to achieve software agents that develop *first-person* meanings, starting with software systems with no such capabilities, either through historical evolution or through individual development.
2. By definition, *first-person* meanings are recognized by their owner.
3. The agent's goals, the description of the agent's capabilities, and the description of properties of the domain are fundamental concepts on top of which the agent may be capable of developing *first-person* meanings.
4. For the purpose of the present research, it is not important to know if the agent is really guided by goals. It is also not important to understand the evolutionary process by which organisms would have become goal-directed individuals. What is actually important is finding a way an agent starts seeing itself as goal-directed.
5. After the agent acquires the knowledge of its assumed goals, and the description of its capabilities, tasks, and environment, it is important for the agent to adopt them, and their relationships, and to shape its future behaviour accordingly.

The next subsections discuss related work, both objections and proposals, regarding the possibility of having computer programs that develop *first-person* meanings in the framework of our research assumptions. In the next section, we examine mainstream work on artificial agents with goals (and other motives). In most cases, the discussed work is not related with the present article. In other cases (i.e., goal recognition), although marginally more related to ours, it does not represent any clear advantage relative to our proposal. Then, we discuss claims about the theoretical impossibility to create computer programs that develop *first-person* meanings. In particular, we analyse the position that only living beings are capable of developing them. Then, we discuss three groups of proposals regarding this problem. The first group addresses the sub problem of symbol reference. The second group stresses the need of a motivation system. Finally, the third group proposes that instead of building computer programs that develop *first-person* meanings, computer scientists should embrace the problem of building software environments from which the desired programs would emerge.

## 2.4 Mainstream Research on Agent Goals

A fundamental notion of our approach is the concept of goal (in the broad sense, as explained in section 2.1). The field of intelligent autonomous agents has produced a vast amount of work related with goals and goal processing. The main purpose of this section is to show that the research related with goals that has been published in the field of autonomous agents is not related to the work presented in this article.

Goals and other motivational states (as desires) play important roles in emotion [27], forgiveness [28], relevance [29], individual regulation process [30], among others. All this work assumes that goals (or the required related motivational concepts) already exist. In our contribution, goals are autonomously discovered and used with a different purpose, namely the generation of *first-person* meanings, and a mechanism by which rule following entities, which do not have goals, may start seeing themselves and the world as if their behaviour were directed by goals.

BDI (Beliefs, Desires and Intentions) accounts of agent behaviour [31] provide models of agent reasoning, planning (for goal achievement) and intention formation [32, 33, 34, 35, 36, 37]. Once again, this is not the main subject of this paper because BDI approaches assume the existence of goals or desires, while we start with agents with no such motivators and propose algorithms that can be used by such rule following entities to start seeing themselves as if they were goal directed agents. However, we also propose that once an agent starts seeing itself as having goals it only becomes a goal directed agent if it adopts the discovered goals. By adopting a goal, we mean shaping its future behaviour in accordance with the discovered goal. Our paper does not focus on this process, nevertheless, we proposed and implemented an agent that changed its behaviour to account for its discovered goals, using a planning algorithm.

In accordance with general theory presented by Conte and Castelfranchi [38], Castelfranchi and colleagues [39, 40] discuss goal-mediated mechanisms of norm compliance, goal delegation, and goal adoption in the context of social action. The major principle is that an agent adopts a goal (or a norm) only if the agent believes that the goal is instrumental for other goals that it already possesses. That is, according to these proposals, it is required that the agent already has goals. In this article, we propose an approach by which rule following entities, which do not have goals, can discover that their behaviour could be explained as if they had goals.

Pezzulo and Castelfranchi [41] propose a mechanism of goal formation: goals are formed as the result of a process that consists of evaluating actual or imagined states of affairs with respect to the agent's drives (or any other reward mechanism). Thus, the formation of a motivator (e.g., a goal, in their case) requires the previous existence of some other motivator (e.g., a drive, in their case). However, in our research, we do not need to assume the previous existence of any kind of motivator. We start with mere rule following entities (computer program, in our case), which are not supposed to have motivators.

In what concerns goal recognition, most of the related work involves open-ended narrative games (namely, the CRYSTAL ISLAND open-world educational game). For instance, Mott, Lee, and Lester [42] (probabilistic models, such as n-gram based models and Bayesian networks), Ha et al. [43] (Markov logic networks), and Min et al. [44] (Long Short-Term Memory-based approach) use information like the narrative state, the actions of the users, and the location of the user in the virtual world to select the users' goals from a previously defined set of possible goals. Sohrabi et al. [45], Pereira et al. [46], Pozanco et al. [47] and Amado et al. [48] propose planning-based goal recognition approaches and assess their proposals on different planning domains, but again the set of possible goals is previously known and used for the determination of the agent's goals.

## 2.5 Theoretical Impossibility of Building Computer Programs with *First-person* Meanings

Although none of them uses the expression *first-person* meanings, for Searle [2] and also for Penrose [3], only exact copies of living beings would be capable of *first-person* meanings because consciousness is generated at the physical level of the brain, in the context provided by the body to which it is coupled.

We argue that this does not have to be the case because we can start by creating a functional architecture [49] with the same properties of the brain coupled to its body, at the desired level of abstraction. Then, it would be possible to create the processes responsible for *first-person* meanings on top of that functional architecture (note that Block [50] presents arguments against functionalism).

Penrose puts forth a more difficult argument: since mathematicians can prove propositions that, according to Gödel [51], cannot be proved by a computer program, there must be processes taking place in the human mind that are not computable. Given that, for Penrose, the only known non-computable physical phenomenon is quantic objective reduction, he hypothesized that consciousness would have a quantic basis. Hameroff and Penrose [52, 53] propose a model of consciousness whose physical basis is the quantic objective reduction taking place in the microtubules of the brain cells, which they consider not computable. Since we require that agents are aware of *first-person* meanings, a manifestation of consciousness, this argument undermines the functionalist approach since it would be impossible to create a computer program capable of exhibiting the desired properties at the quantic level because they were assumed to be non-computable.

LaForte, Hayes, and Ford [54], and Krajewski [55] show that Penrose's conclusion that the human mind involves non-computable processes is wrong. LaForte, Hayes, and Ford show that the mathematicians that have proved the sentences that are not provable by an algorithm could only do that because they did not prove the soundness of their proofing procedure, which would be required for a proper proof. According to these authors, proving the soundness of the used proof procedure is impossible, even for those mathematicians. Krajewski presents and proves a general theorem, which shows that every kind of argument similar to that of Penrose must be either circular or lead to an inconsistency.

Ziemke [4], Froese and Ziemke [5], and Sharkey and Ziemke [56] claim that any externally developed mechanisms aimed at creating meanings for a computer program will always produce arbitrary results for the computer program itself exactly because they are externally created.

To overcome this obstacle the computer program would have to create and maintain, from within, its whole network of parts and processes, as it is the case with living beings. This way, the meaning development mechanism of the program, being developed and maintained by the program itself, would produce non-arbitrary meanings for the program.

Esteves and Botelho [57] presented an approach, according to which, the computer program develops and maintains all of its parts and processes from within, starting from an initial pre-programmed zygote. Since this initial zygote is pre-programmed, the approach continues to face objections because the initial zygote of living beings is the result of the co-evolution of the species and its environment; it was not externally created.

We believe that the fact that a living being develops itself and all its parts and processes from within does not constitute an advantage concerning the possibility of having *first-person* meanings. Some of the requirements for a system to develop *first-person* meanings are *(i)* being capable of determining the importance of objects, events, and situations in face of their goals, their capabilities, and the characteristics of the domain; and *(ii)* being aware of (recognizing) that importance. A system can do this if it sees itself as if it were a constrained goal-directed entity in relation with its environment, considering that "seeing itself in a certain way" is being aware. Simple living beings and computer programs, as other rule-following systems, do not in fact have goals; they just do what the forces applicable to their inner workings impose them given what happens to them. However, once equipped with appropriate processes, a rule-following system may see itself as goal-directed.

Besides, it is not absolutely accurate that a living being creates all its processes from within. In fact, living beings whose behaviour is totally determined by natural forces (e.g., electrical and gravitational forces) do not create those forces. Therefore, the lowest level processes taking place in a living being, those at the physical level, are determined by forces external to the living being.

## 2.6 Simple Organisms and *First-person* Meanings

Thompson and Stapleton [24], and Froese and Ziemke [5] claim that the life of even the simplest organism can be described as a sense-making activity. For instance, a specific motile bacterium moves along the gradient of sugar concentration towards its highest value, as a result of its metabolism. Even simple organisms, as the mentioned bacterium, regulate their interactions with the world in such a way that transforms the world into a place of salience, meaning, and value. According to these authors, this shows how the relation of even simple organisms with their environment becomes meaningful as a result of the organisms' metabolism.

Thompson [58] and Froese [59] recognize that this kind of meaning (sense-making) is not to be equated with the *first-person* perspective of conscious beings. However, Froese and Ziemke [5] believe that it is impossible to have systems with *first-person* meanings that are not rooted on the lower-level sense-making mechanisms resulting of their metabolism.

We argue that the meaningfulness that is created by simple organisms, such as bacteria, as a result of their metabolism, is in the head of an external observer, exactly in the same way that critics of artificial intelligence have identified in the meaningfulness of the behaviour of artificial systems. In fact, the behaviour of simple organisms results, in each moment, of the interplay between the natural forces applicable to their inner workings and their state (internal and external), at that moment. In short, simple organisms are just rule-following systems, exactly in the same way as computer programs. None of them has a *first-person* perspective of what they do and of their environment. One of the requirements for the emergence of a *first-person* perspective in a system is the capability of relating what it does and what happens to it with its assumed goals, its capabilities, and the constraints and opportunities its environment places on the possibility of the system to achieve its goals. However, goals and goal-based explanations may be mere useful abstractions that enable observers to better understand the behaviour of complex systems [9]. For instance, a thermostat may be described as if it had the goal of maintaining the temperature within a specified range. However, the behaviour of the thermostat is not determined by goals; it is determined by applicable low-level forces, for instance, the instructions of a control program.

How then is a system to develop a *first-person* perspective, if it does not have any goals or other motives? Only systems capable of rationalizing (explaining) their behaviour and their environment as if they were constrained goal-directed entities can have a *first-person* perspective.

This is the reason we claim that, regarding the development of *first-person* meanings, it is not important to know whether or not the system actually has goals; what is important is seeing itself as a goal-directed entity.

Given that the sense-making of simple organisms is not an intrinsic property of those organisms but, instead, lies in the mind of the observer, and hence metabolism is not a requisite for *first-person* meanings, we believe that it is possible to individually develop or to historically evolve computer programs with no *first-person* meanings into computer programs with *first-person* meanings. The mentioned development or evolution process must, among other things, provide the resulting computer programs with the capability to see themselves as constrained goal-directed entities, and to become aware of the importance of things, events, and situations relative to the program's goals, capabilities, and to the properties of its environment.

## 2.7 The Sub Problem of Symbol Reference

Although not the main focus of this article, which focuses on the importance of events and states of affairs to a system in relation to the system's goals, the general problem of *first-person* meanings includes also the symbol reference problem. We briefly describe some of the approaches to the symbol reference problem.

Several authors make use of the physical components of robots, e.g., sensors and actuators, to tackle this problem. Namely, Harnad [1], and Cangelosi, Greco, and Harnad [17] ground the symbols of their systems on the system's sensors, while Mugan and Kuipers [60] ground the plans of their robots on the motor commands of their actuators. Although relevant, this kind of work does not equip the agent with the awareness of the importance of objects, situations, and events to the agent's goals.

Another approach to the symbol reference problem relies on the interactions between systems. For example, Vogt [18], Steels [20, 21], and Machado and Botelho [19] involve their systems in especially designed social interactions through which they develop, from scratch, a shared lexicon of symbols, each of which is connected to the object (and sometimes the category) they represent. Once again, symbol reference is not the same as its importance.

Pezzulo and Castelfranchi [41] explain the way symbols may become detached from their original sensorimotor interaction, without losing their aboutness, their relevance, and their normativity.

A different approach is to look at the symbol reference problem from the behavioural perspective. For example, Anderson and Perlis [10] claim that computational tokens have or acquire representational powers for a computer program if they play consistent behaviour-guiding roles for the program. In this case, the meaning of each symbol is rooted in its behaviour-guiding role, that is, in its causal powers. Even though the behaviour of a computer program is consistently dependent on each symbol (due to the assumed consistent behaviour-guiding powers of the symbol), the program would not be capable of stating the meaning of each of its symbols.

For Brooks [61], symbols and symbolic processing are not required for intelligent behaviour therefore his robots avoid the symbol reference problem. In our point of view, if the robot is to determine the importance of objects, events, and situations, it must see all it does and all that happens to it in the light of its goals. Moreover, as Vogt [18] argues, several high-level mental activities, such as language, require symbols (but see the work of Cuffari, di Paolo, and de Jaegher [62] for a contrasting opinion).

## 2.8 The Need of a Motivational System

For Dreyfus [20], systems may develop *first-person* meanings only if they have body with body-related needs, which give rise to motivated behaviour. Steels [63] and Birk [64] developed robots that successfully learn to stay alive in their environment relying only on mechanisms aimed at preserving internal variables within specified values, which may be seen as a form of motivated behaviour. Savage [65] discusses a set of processes by which complex motivated behaviour can be developed in artificial agents, either by gradual improvement through the agent's experience or by historical evolution.

We fully agree that a motivational system is required for *first-person* meanings. However, we argue that, even if the presented systems [63, 64] act as if they knew what is important to them, they do not really have any idea of the importance of objects, events, and situations, in the sense that they would not be capable of saying that they are important (the systems would not recognize that importance). *First-person* meanings is the recognition, by the system, of such importance.

## 2.9 Emergence of Agents with *First-person* Meanings

As already stated before, Froese and Ziemke [5], among others, contend that it is impossible to build computer programs with *first-person* meanings because the computer program would have to create and maintain itself and the whole network of its parts and processes from within. To circumvent this problem, Froese and Ziemke [5] present the quite radical idea that instead of trying to build programs with *first-person* meanings, scientists should focus on the definition and creation of environments with such dynamics that enable the emergence of computer programs with *first-person* meanings.

We feel sympathetic towards this idea, which we thoroughly considered adopting and implementing. Although we have not yet excluded this possibility, we have decided to tread a more traditional route for the following reason. Since we know of no literature reporting the emergence, from currently existing environments, of computer programs with *first-person* meanings, we think that currently existing environments will not do. Therefore, we would need to create new and different environments from which the desired computer programs would emerge. However, we cannot be sure that a computer program with *first-person* meanings emerging from the new environment would be capable of tackling the problems of currently existing environments. Chances are that the emerging program would contribute to our understanding of the problem but not to solve the problems of currently existing environments for which we already need better programs. Besides, we feel that the problem of defining environments from which the desired computer programs would emerge is even less understood than the problem of defining the desired programs in the first place.

## 3   Research Approach, Proposals and Achievements

Following the research assumptions presented in the previous section, our research agenda consists of investigating the following problems:

1. Mechanisms by which a computer program *(i)* discovers that its behaviour is (or could be) goal-directed, *(ii)* discovers the description of its capabilities, and *(iii)* discovers properties of the task it performs and the environment it inhabits;
2. Mechanisms by which a computer program may use its knowledge about (cold awareness of) its goals, its capabilities, and the properties of the task it performs and of the environment it inhabits to create cold *first-person* meanings of objects, events and situations;

3. Mechanisms by which a computer program may use its knowledge about its goals, about its capabilities, and about the properties of the task it performs and of the environment it inhabits to shape its future behaviour;

4. Mechanisms by which phenomenal experience may be developed in computer programs. We are interested, in particular, in the phenomenal awareness of the importance of objects, events, and situations in relation to the agent's goals, its capabilities, and the constraints and opportunities presented by the environment, which is the phenomenal component of *first-person* meanings, in the sense we have chosen to address.

5. Everything we do in engineering must be validated therefore all the described problems must be associated with adequate validation procedures. However, designing formal validation procedures for some of our targeted research goals, in particular feelings in the phenomenological sense, will be a major challenge. The design of such validation procedures is another problem of the proposed approach. However, we believe that designing such tests will only be possible after we have gained a deeper understanding of the concepts we are dealing with.

This article contributes mainly to the two first problems of the presented research agenda. A smaller contribution to the third problem is also presented.

In our approach there is no single concept that can be equated with (cold) *first-person* meanings. On the contrary: almost every concept acquired by the agent from the observation of its behaviour (see tables 1, 2, and 3) reflects some aspect of *first-person* meanings, is used in the definition of other concepts that reflect some manifestation of *first-person* meanings, or both. Section 3.1 presents rigorous formalizations of all concepts learned by the agent, using the approach we have proposed. Given that the main technical contribution of this paper is the set of algorithms and derivation schema used to acquire the *first-person* meanings related concepts presented in tables 1, 2, and 3, our formalization was designed from the perspective of enabling the acquisition / learning of some of these concepts and the derivation of the others. Given its main focus, the article does not present a formalization aimed at enabling agents to use the acquired concepts in reasoning, decision-making, and action selection.

Besides the mentioned formalization, section 3.1 also discusses the ways in which the concepts acquired by the agent reflect some manifestations of *first-person* meanings. For instance, the predicates Achieved/3 and Contributed/3 allow the agent to understand its behaviour as a meaningful endeavour to reach its goals. Achieved(State, Action, RelevantProps) means that the specified action, executed in the specified state achieved the specified relevant effects. A proposition is a relevant effect for the agent (from a *first-person* perspective) if it is one of the agent's goals (also learned) or if it is the precondition of an action executed by the agent in a future state that actually "consumed" the proposition, which is captured by the predicate Contributed/3. All this is also part of the *first-person* understanding of the goal directed nature of the agent's behaviour.

However, these are not the only learned predicates capturing aspects of *first-person* meanings. For instance, the fact that the agent is constrained by its capabilities and by the structure of the domain is captured by the predicates and functions describing the preconditions, effects, validity conditions of the agent's actions, and the precedence relations between the states of the agent's behaviour (the also learned predicate MustPrecede/2).

The first subsection of the present section describes the concepts that have been acquired by an initial program with no understanding of what it does and of what happens to it, and the domain independent algorithms that have been used for the acquisition of those concepts.

The second subsection describes a proof of concept scenario and shows that the achieved results encourage furthering our research.

## 3.1 Understanding what Happens

The conceptual framework of our work is represented in Fig 1. The software agent is situated in its environment through its sensors and actuators. In addition to the sensors and actuators, the software agent has three additional components: the executor, the observer, and the adopter.

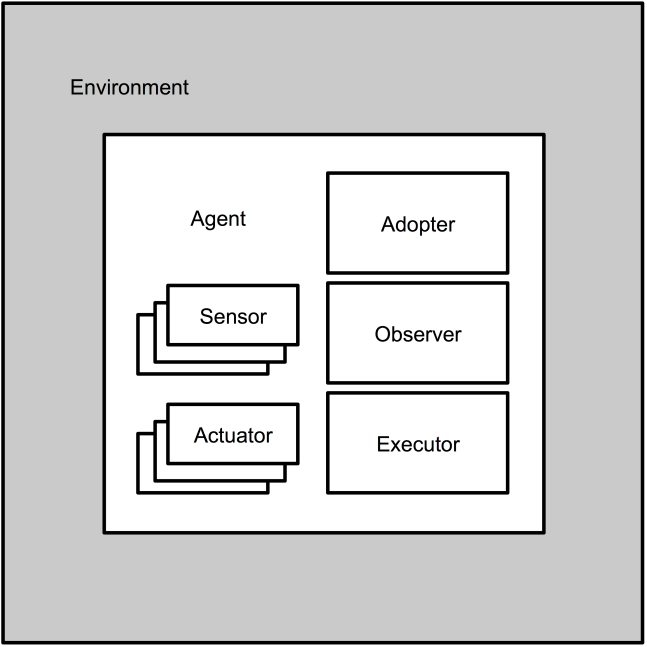

**Fig 1 Functional conceptual framework**

All of these components share the same sensors and actuators. Besides, the observer component has direct access to the actions executed by the executor component, most in the same fashion as the internal observation function of the brain (the "inner eye"), as proposed by Singer [14]). The executor component is programmed to execute a given task with no sense of goals, with no explicit knowledge describing the agent capabilities (sensors and actuators), and with no explicit representations of the domain. It just senses the environment, through the agent's sensors, decides what to do (this does not have to involve any deliberative process relying on explicit knowledge representations), and acts on it through the agent's actuators. The observer component senses the environment exactly as sensed by the executor component, and also senses the actions being performed by the executor. The observer is responsible for learning the agent's goals, the description of the agent's capabilities, and several properties of the agent domain. The adopter uses the knowledge acquired by the observer to shape the agent's future behaviour and to produce conscious feelings of what happens, that is, phenomenal *first-person* meanings. This conceptual framework is described at the functional level. That is, the different agent components correspond to agent functions, not necessarily to different structural components.

The main focus of the work reported hereunder is the observer functional component and a small part of the adopter functional component. We have not addressed the generation of conscious feelings, in the phenomenological sense.

In what follows, a behavioural instance is the sequence of actions the agent's executor component performs. A behavioural class is the set of behavioural instances with the same purpose, i.e., the set of behaviours the agent executes for alternative configurations of the same task the agent's executor was programmed to perform.

In the remainder of the section, a behavioural instance is described as an initial state followed by a sequence of action-state pairs. The state of the last action-state pair of the sequence is called the final state. As the agent's executor executes the actions of a behavioural instance, the agent's observer senses the states and the executed actions.

We have defined a set of concepts to be learned by the agent's observer component as it observes what is being done by the agent's executor component. Some of the defined descriptions are properties of the observed entities of the domain (Table 1, e.g., Proposition/1, Action/1 and Goal/1). The other descriptions relate entities of the domain (Table 3 and Table 2, e.g., Achieved/3, Contributed/2, MustPrecede/2, Precond/1). Domain entities comprise actions and propositions. A state is a set of positive atomic propositions representing their conjunction. For convenience, states are identified by unique identifiers. We have chosen to use non-negative integers to index states. State $S_0$ of a behavioural instance is its initial state. States are numbered by increasing order of their occurrence.

Descriptions are organized into two other groups according to a different criterion: those that apply only to specific behavioural instances (Table 3, e.g., Achieved/3 and Contributed/2) and those that are independent of specific behavioural instances (Table 2, e.g., MustPrecede/2 and Precond/1).

Finally, some descriptions are represented by functions (e.g., PosEffects/1 and NegEffects/1) whereas others are represented by predicates (e.g., Achieved/3, MustPrecede/2).

**Table 1. Learned entity properties**

| Proposition(*Entity*) | *Entity* is a proposition |
|---|---|
| StaticProposition(*Entity*) | *Entity* is a static proposition (true in all states of all behavioural instances of the same class). |
| FluentProposition(*Entity*) | *Entity* is a fluent proposition (true only in some states) |
| Action(*Entity*) | *Entity* is an action. |
| Goal(*PropsSet*) | The goal of the agent is the conjunction of all propositions of the set *PropsSet*. That is, all the agent behavioural instances of the same behavioural class are directed to achieve a state in which all propositions in *PropsSet* are true. |
| Desired(*Prop*) | *Prop* is a desired proposition. A proposition is desired if it is one of the agent's goals. |
| Mandatory(*Prop*) | *Prop* is a mandatory proposition in the sense that, although it is not a goal, there has to be at least one state in the agent behaviour, in which the proposition is true. The obligation results of the interaction of the task structure, the agent's goals and the agent's capabilities. |

**Table 2. Descriptions that relate entities. Applicable to all behaviours**

| | |
|---|---|
| MustPrecede($Prop_1$, $Prop_2$) | In all instances of the same class of behaviours in which $Prop_1$ and $Prop_2$ have occurred, $Prop_1$ and $Prop_2$ occur both in the initial state or $Prop_1$ must occur before $Prop_2$ occurs. The precedence relation arises exclusively of the task structure, the agent's goals and capabilities. It does not arise of the fact of $Prop_1$ being a precondition of any of the actions that lead from the state in which $Prop_1$ is true to the state in which $Prop_2$ is true. |
| Precond(*Act*) | *Precond(Act)* is the set of preconditions of the action *Act*, in the sense that *Act* can only be executed in a state in which all of its preconditions are true. |
| PosEffects(*Act*) | *PosEffects(Act)* is the set of positive effects of *Act*, that is, the set of propositions that become true in the state immediately after *Act* is executed. |
| NegEffects(*Act*) | *NegEffects(Act)* is the set of negative effects of *Act*, that is, the set of propositions that cease to be true in the state immediately after *Act* is executed. |
| ValidityCondition(*Act*) | *ValidityCondition(Act)* is the validity condition of the action *Act*. The validity condition of an action specifies the set of valid instantiations that can be applied to the action variables. |

**Table 3. Descriptions that relate entities. Applicable to specific behaviours**

| Achieved(*State*, *Act*, *PropsSet*) | The propositions in the set *PropsSet* constitute the relevant effects of the action *Act*, executed in the state *State*. Relevant in the sense that each of them is either one of the agent's goals or one of the preconditions of an action that actually used it in a future state of the same behaviour. |
|---|---|
| Contributed(*State*, *Prop*, *Act*) | Proposition *Prop*, true in the state *State*, contributed to the execution of the action *Act* in the state *State*, because it is one of *Act*'s preconditions. |

Of the mentioned learned concepts, the predicates Goal and Desired capture the agent's goals; the predicate Action and the functions Precond, PosEffects, NegEffects, and ValidityCondition describe the agent's capabilities; and the predicates MustPrecede and Mandatory represent properties of the agent task (considering its goals and capabilities). These last two also capture cold *first-person* or intrinsic meanings of specified properties of the environment to the agent because, if the agent is to fulfil its goals, it knows it must satisfy these two classes of constraints (for all behavioural instances of the same class). All the above concepts represent cold *first-person* meanings or are used to define the predicates Contributed and Achieved. These latter predicates also represent cold *first-person* meanings of specified properties of the environment observed by the agent and of specific actions performed by the agent in the particular considered behaviour.

Some of the presented concepts can be formally defined (e.g., fluent propositions). In such cases, we present the axioms that can be used for their deduction. Some other concepts are not formally defined (e.g., goals). We have defined and implemented totally domain independent algorithms that were used by the agent's observer component to learn all the described concepts. In the remainder of this section, we present the fundamental ideas behind each of those algorithms.

**Propositions and actions**. Propositions and actions are the output of the agent's sensors. A state is exactly the set of propositions observed by the agent's sensors at a given instant. For convenience, the State/1 predicate holds state identifiers instead of state propositions. The function StateProps/1, applied to a state identifier, returns the set of propositions true in the state identified by the specified state identifier.

The predicate Proposition/1 is determined through the set union of all states of all behavioural instances of the agent. The predicate StaticProposition/1 is determined through the set intersection of all states of all behavioural instances of the agent.

Knowing the set of all propositions and the set of all static propositions, it is possible to determine the fluent propositions using the following axiom schema:

$$(\text{Proposition}(P) \land \neg \text{StaticProposition}(P)) \Rightarrow \text{FluentProposition}(P)$$

That is, the set of fluent propositions is the difference between the set of all propositions and the set of the static propositions. Although implementation details are not important, all the algorithms have been implemented in the Prolog language. The implementation of all presented deduction axiom schemata is remarkably similar to the actual axioms. This is illustrated by the following example corresponding to the axiom schema just presented.

```
fluent_proposition(P) :- proposition(P), not static_proposition(P).
```

The predicate Action/1 is computed through the set union of the actions observed in all behavioural instances of the agent.

**Agent goals: Goal/1**. The agent's goals, learned through the observation of its behaviour, are represented by the predicate Goal/1. Goal(*PropsSet*) means the agent behaves as if its goal were the conjunction of all propositions in the set *PropSet*.

Whether or not the agent really has goals, the purpose of this research is to provide the means for the agent to understand its relation with the environment as if its behaviour were guided by goals. We have assumed that, in case of success, the agent behaviour ends when the agent's goals are achieved. This assumption is the basis for the proposed goal discovery algorithm. The set of agent's goals is determined through the intersection of the sets of fluent propositions of the final states of all the observed behavioural instances of the same class. We have provided an example of the implementation of an axiom schema (for the deduction of fluent propositions). However, not all algorithms correspond to logical axioms; several are learning algorithms that must process all instances of the same class of behaviours. In general, those learning algorithms perform set operations with the entities of the domain. The following pseudo-code exemplifies that class of algorithms for the case of determining the goals of the agent. This time, we do not provide the Prolog code because procedural Prolog code is not straightforward to understand for non-Prolog programmers.

All such learning procedures load the descriptions of the behaviours from files where they have been kept.

% Procedure that computes the set of the agent's goals from a list of behavioural instances

% List_of_behaviours: observed behavioural instances, each recorded in a file

% In the end, the fact goal(UpdatedGoals) will contain the agent's goals.

**compute_goals**(*List_of_behaviours*) {

    **foreach** *Behaviour* **in** *List_of_behaviours* {

        load_beahvior(*Behaviour*) % Read the behaviour from its file and load it into the program

        % Fluent propositions of last state of current behavioural instance

        *FluentProps* = last_state_fluent_props()

        update_goals(*FluentProps*)

    }

}

% Update the set of goals with the newly received goals (NewGoals)

% Updated goals are the result of the set intersection of the previous set of goals

% with the new goals

**update_goals**(*NewGoals*) {

    **if** goals do not exist yet **then** store goal(*NewGoals*)

    **else** {

        **remove** goal(*PreviousGoals*) % *PreviousGoals* becomes known

        *UpdatedGoals* = setintersection(*PreviousGoals*, *NewGoals*)

        **store** goal(*UpdatedGoals*)

    }

}

Upon completion, the fact goal(*Goals*) will represent the set of the agent's goals.

It is worth noting that the presented pseudo-code does not directly correspond to the actual implementation. The actual program does not implement a separate process for computing goals; it performs several computations at once, for instance the set of observed propositions, the set of observed actions, and the set of goals, among others.

Mainly for convenience, we propose the definition of a desired proposition as a proposition that is one of the agent's goals. This definition leads to the following axiom schema.

$$(\text{Goal}(PSet) \land P \in PSet) \Rightarrow \text{Desired}(P)$$

The algorithm used for determining desired propositions just needs to have access to the set of the agent's goals.

**Mandatory propositions: Mandatory/1**. We say that if a proposition $P$ must precede another proposition and $P$ is not one of the agent's goals then P is mandatory. We believe it would be possible to identify other conditions under which a proposition should be considered mandatory but, for the moment, we will stick to this one.

Knowing that a proposition is mandatory is meaningful to the agent, given that it knows that it will only achieve its goals if its behaviour includes a state in which the mandatory proposition holds.

The following axiom schema captures the proposed intuition.

$$(\exists q\, \text{MustPrecede}(P, q) \land \neg \text{Desired}(P)) \Rightarrow \text{Mandatory}(P)$$

As it will be seen later, MustPrecede($Prop_1$, $Prop_2$) means that, due to the way the world works, to the task structure, and to the agent's goals and capabilities, if propositions $Prop_1$ and $Prop_2$ occur in a behavioural instance then either they both occur in the initial state or $Prop_1$ must occur before $Prop_2$.

Given that MustPrecede/2 always relates fluent propositions, the algorithm for computing the predicate Mandatory/1 restricts the set of all fluent propositions to those that are not desired but must precede another proposition.

**Precedence relations: MustPrecede($Prop_1$, $Prop_2$)**. As it has been defined, a precedence relation is intrinsically meaningful to the agent because it knows it cannot achieve its goals if it does not satisfy the learned precedence.

We start by the definition of the precedence relation for a single behavioural instance. Then we provide the general definition, valid for all behavioural instances of the same class.

Before the definition, a note about transitivity of the precedence relation, as it is used in our research, is in place. Contrarily to the usual precedence concept, the precedence relation, as we have conceptualized it, is not transitive. It is possible to have MustPrecede ($P_1$, $P_2$) and MustPrecede($P_2$, $P_3$) without MustPrecede($P_1$, $P_3$). Consider the example of the following two behavioural instances, $B_1$ and $B_2$ in Fig 2:

| Behavior $B_1$ | | Behavior $B_2$ | |
|---|---|---|---|
| State | State Propositions | State | State Propositions |
| $S_0$ | $P_1$ | $S_0$ | $P_2$ |
| $S_1$ | $P_2$ | $S_1$ | $P_3$ |
| $S_2$ | $P_3$ | $S_2$ | $P_1$ |
| | | $S_3$ | $P_2$ |
| Precedence relationships: $<P_1, P_2>$ $<P_1, P_3>$ $<P_2, P_3>$ | | Precedence relationships: $<P_1, P_2>$ $<P_2, P_3>$ $<P_2, P_1>$ $<P_3, P_1>$ $<P_3, P_2>$ | |

**Fig 2 Precedence relation in two examples**

The precedence relation shared by the two behaviours $B_1$ and $B_2$ consists only of the pairs $<P_1, P_2>$ and $<P_2, P_3>$, but not by the pair $<P_1, P_3>$. If we had defined the precedence relation as being transitive, one of the pairs $<P_1, P_2>$ or $<P_2, P_3>$ would have to be excluded. The major problem then would have been to decide which of them to exclude. Instead of finding a criterion to exclude one of these pairs, we decided to define the precedence relation as non-transitive.

<u>Precedence in a singular behavioural instance</u>: Precedes($P_1$, $P_2$) holds for a given instance of a behavioural class if the fluent propositions $P_1$ and $P_2$ are both true in that behaviour's initial state; or if $P_1$ occurs before $P_2$, and $P_1$ is not a precondition of any of the actions of the action sequence that leads from the state in which $P_1$ is true (but not $P_2$) to the state in which $P_2$ is true.

That is, if $P_1$ and $P_2$ are not true in the initial state, there must be a sequence of actions, *ActSequence*, such that (*i*) performing *ActSequence* starting in the state $S_1$ leads to the state $S_2$; (*ii*) $P_1$ occurs in $S_1$, and $P_2$ occurs in $S_2$ but not in $S_1$ or in any of the states between $S_1$ and $S_2$, if any; and (*iii*) $P_1$ is not a member of the set of preconditions of any of the actions in *ActSequence*.

Precedence relation, defined in a class of behaviours: MustPrecede($P_1$, $P_2$) holds for a behavioural class, iff, for each instance of that behavioural class, *b*, in which both $P_1$ and $P_2$ occur, $P_1$ precedes $P_2$ in that behavioural instance, *b*: Precedes($P_1$, $P_2$). To avoid situations in which the precedence relation is satisfied exclusively by behavioural instances in which $P_1$ and $P_2$ occur simultaneously in the initial state, we additionally impose that there must be at least one behavioural instance in which $P_1$ actually precedes $P_2$.

The algorithm that learns the precedence relation by observation closely follows the described definition. First, we defined an algorithm that discovers the precedence relation in a single instance of the behavioural class. Then, we defined the algorithm that checks if the precedence relation holds in all behavioural instances and if there is at least one of them in which $P_1$ actually precedes $P_2$. This second component of the algorithm calls the single instance component, discarding those behavioural instances in which $P_1$ and $P_2$ do not occur at all.

**Action preconditions: Precond(*Act*)**. Precond/1 is a function such that Precond(*Act*) represents the set of all preconditions of action *Act*. Action preconditions are fluent propositions that must be true for the action to be executed.

The algorithm to determine the preconditions of an action performs the intersection of the sets of fluent propositions of all states of all behavioural instances in which the action is executed.

Although this idea is correct, the fact that some actions are performed only too few times, even if all instances of the same behavioural class are considered, may lead to sets of preconditions with more propositions than those actually necessary for the action to be executed.

To circumvent this problem, it was necessary to generalize actions instead of considering only the observed instances of the same action. For example, instead of considering every instance of a given action (such as the blocks world action Move(A, $P_1$, B) - moving block A from position $P_1$ to the top of block *B*), it is necessary to consider the abstract action (Move(*block*, *from*, *to*), in which *block*, *from* and *to* are variables). Using generalization of the action arguments allowed the algorithm to discover the correct set of action preconditions.

In addition to solving the mentioned problem, caused by action instances that are seldom executed, generalization enables to use the abstract action preconditions for problems that require action instances that were never observed, as it is the case in planning problems. This second reason led us to use generalization also for determining the action effects and the action validity conditions.

**Action effects: PosEffects(*Act*) and NegEffects(*Act*)**. PosEffects/1 and NegEffects/1 are functions that return action effects. PosEffects(*Act*) represents the set of the positive effects of the action *Act*; and NegEffects(*Act*) represents the set of its negative effects.

The action's positive effects are the propositions that become true immediately after the action is executed. The action's negative effects are those propositions that were true before the action is executed and cease to be true immediately after the action is executed.

The set of positive effects of an instance of an action is determined as the difference between the set of propositions of the state observed immediately after the action instance has been executed and the set of propositions holding in the state in which the action instance was executed.

The set of negative effects of an instance of an action is obtained through the difference between the set of propositions of the state in which the action instance was executed and the set of propositions of the state immediately after the execution.

After determining the effects of all action instances, the algorithm performs generalization on the action arguments, leading to the general expression of the effects of all observed actions.

**Action validity condition: ValidityCondition(*Act*)**. The functional expression ValidityCondition(*Act*) represents the conditions that must be satisfied by variables of the action *Act* that make it a valid action. That is, it specifies the valid instantiations of those variables. For instance, the validity condition of the blocks world action Move(*block*, *from*, *to*) is that the variable *block* must be a block, the variables *from* and *to* must be places where a block can be placed, and all the variables must have different values:

$$(\text{Block}(block) \land \text{Place}(from) \land \text{Place}(to) \land block \neq from \land block \neq to \land from \neq to)$$

We emphasize that, as it was the case with preconditions and effects, it is necessary, for certain problems such as planning, that the action validity condition is generalized, i.e., it must specify a set of values for the variables that is more general than just the observed instantiations.

The main idea underlying the algorithm used for determining a generalized validity condition of an action consists of finding the static predicates that cover all the observed instances of the same action. First the algorithm tries to apply each one-place predicate to each of the action variables and checks if all observed instances of the action are covered by the tried predicates. Then, the algorithm tries to apply all binary static predicates to all ordered pairs of variables. Then it moves to the ternary predicates, and so on, until all static predicates have been tried.

Each time the algorithm tries to use an N-ary static predicate with all N-ary tuples of variables, it uses first the predicates with a smaller extension as a way of avoiding overgeneralizations, which would have a greater likability of being wrong.

**Description of specific behaviours: Achieved/3 and Contributed/3**. Achieved/3 and Contributed/3 do not reflect general properties of actions or propositions. They reflect the usefulness, the purpose, of specific actions and propositions of specific behaviours.

The relationships Achieved(*State*, *Act*, *PropsSet*) and Contributed(*State*, *Prop*, *Act*) represent the cold *first-person* meanings of actions and propositions of specific behaviours to the agent.

Achieved(*State*, *Act*, *PropsSet*) expresses the agent's knowledge of the reasons justifying the execution of a given action in a specific behaviour: achieving one of its goals or as a means to achieve a state in which another action could be and was actually executed. The propositions in the set *PropsSet* constitute the relevant effects of the action *Act*, executed in the state *State* of the specific observed behaviour. Action effect, in the sense that the action was executed in a state in which no such propositions were true leading to a state in which they all become true. Relevant in the sense that each of them is either one of the agent's goals or one of the preconditions of an action that actually used it in a future state of the same behaviour.

Contributed(*State*, *Prop*, *Act*) expresses the agent's knowledge of the usefulness of a given proposition, true in a specific state of a specific behaviour: being the precondition of an action actually executed at that state. Proposition *Prop*, true in state *State* of the specific observed behaviour, contributed to the execution of action *Act* in the same state of the same behaviour, because it is one of *Act*'s preconditions.

For the formalization of the axiom schema that can be used for the deduction of Contributed/3 relationships, it is necessary to use the predicate NextState/3 that has not yet been introduced. NextState($S_1$, $A$, $S_2$) relates a state identifier ($S_1$), the action that is performed in that state ($A$), and the identifier of the new state resulting of the action execution ($S_2$). NextState/3, as well as State/1 and StateProps/1, refers to states, actions, and propositions actually observed by the observer component of the agent. For instance, $\exists t$ NextState($S$, $A$, $t$) means that *(i)* there is a state in the considered observed behaviour, identified by $S$, and that *(ii)* the action $A$ was actually executed in that state, leading to another state whose identification is not specified. These predicates and function do not refer to possibilities; they refer to actually executed behaviour.

( State($S$) ∧ $P \in$ StateProps($S$) ∧ Action($A$) ∧

$\exists r$ NextState($S$, $A$, $r$) ∧ $P \in$ Precond($A$) ) $\Rightarrow$ Contributed($S$, $P$, $A$)

The positive effect $P$ of action $A_1$ is relevant if $P$ is one of the agent's goals or if $P$ is one of the preconditions of a second action $A_2$ that could be executed after $A_1$ because that very occurrence of $P$ was true when $A_2$ was executed. If $A_2$ has been executed in a state occurring after the one that ensues immediately after the execution of $A_1$, $P$ must be true in all states, from $S$ until the state in which $A_2$ was executed.

The axiom schemata for the deduction of the relevant effects of an action requires the definition of the relation PathTrue/3 such that PathTrue($S_i$, $S_j$, $P$) means that $P$ is true in all states from $S_i$ to $S_j$ inclusive.

(State($S$) ∧ State($T$) ∧ $S \leq T$ ∧ $\neg \exists r$ (State($r$) ∧ $S \leq r \leq T$ ∧ $P \notin$ StateProps($r$)))

$\Rightarrow$ PathTrue($S$, $T$, $P$)

The relation ≤, between state identifiers, reflects a temporal ordering among states. Being *S* and *T* two state identifiers, S ≤ T means that the state identified by *S* occurred before the state identified by *T*, or that they identify the same state.

( State(*S*) ∧ Action(*A*) ∧ ∃*r* NextState(*S*, *A*, *r*) ∧

Desired(*P*) ∧ *P*∈PosEffects(*A*)) ⇒ RelevantEffect(*S*, *A*, *P*)

∃*r* ( State(*S*) ∧ Action(*A*) ∧ NextState(*S*, *A*, *r*) ∧ *P*∈PosEffects(*A*) ∧

∃*t* (State(*t*) ∧ *S* ≤ *t* ∧ PathTrue(*r*, *t*, *P*) ∧ ∃*a* Contributed(*t*, *P*, *a*)))

⇒ RelevantEffect(*S*, *A*, *P*)

*PropsSet*, in the relationship Achieved(*S*, *A*, *PropsSet*) represents the set of the relevant effects of the action *A*, executed in the state *S*. The following axiom schema captures the definition:

Achieved(*S*, *A*, {*p*: RelevantEffect(*S*, *A*, *p*)})

The algorithms for determining the relations Contributed/3 and Achieved/3 apply the axiom schemata presented for their deduction, for which they need the following relations and functions: State/1, Action/1, Desired/1, StateProps/1, NextState/3, PosEffects/1 and Precond/1. Determining Achieved/3 also requires the relation Contributed/3 (or else, its definition).

For determining the relation Contributed/3 the algorithm computes, for all states of the considered behaviour, the set of tuples <*S*, *P*, *A*> such that the proposition *P*, true in the state *S*, is one of the preconditions of the action *A*, executed in state *S*.

For determining the relevant effects of an action (i.e., relation RelevantEffects/3), the algorithm computes all tuples <*S*, *A*, *P*> such that the action *A*, executed in the state *S*,

(i) generated at least one of the agent's goals (*P*); or

(ii) generated one of the preconditions *P* of another action (*A'*) executed in a future state; *P* was true in all states from the one immediately after the execution of *A* until the one in which *A'* was executed.

These tuples are then used to generate the tuples of Achieved/3 relation, which include, in their third element, the set of all relevant effects of *A*, executed in the state *S*.

## 3.2 Proof of Concept and Discussion

The described definitions and algorithms were demonstrated in a simple scenario of the blocks world. Given their independence of the domain, the presented definitions and algorithms could have been demonstrated in other scenarios.

**Blocks world scenario**. In this version of the blocks world (Fig 3), the agent moves blocks around until they are all stacked on top of each other. The place where the blocks will be stacked is irrelevant. There are always three equal blocks – A, B, and C – and four spaces on a table – $P_1$, $P_2$, $P_3$, and $P_4$. Each block may be placed on one of the table positions or on top of any other block. The predicate *Place/1* is used to hold the places where blocks may be located. The final stack must contain the block B on top of the block C and the block A on top of the block B.

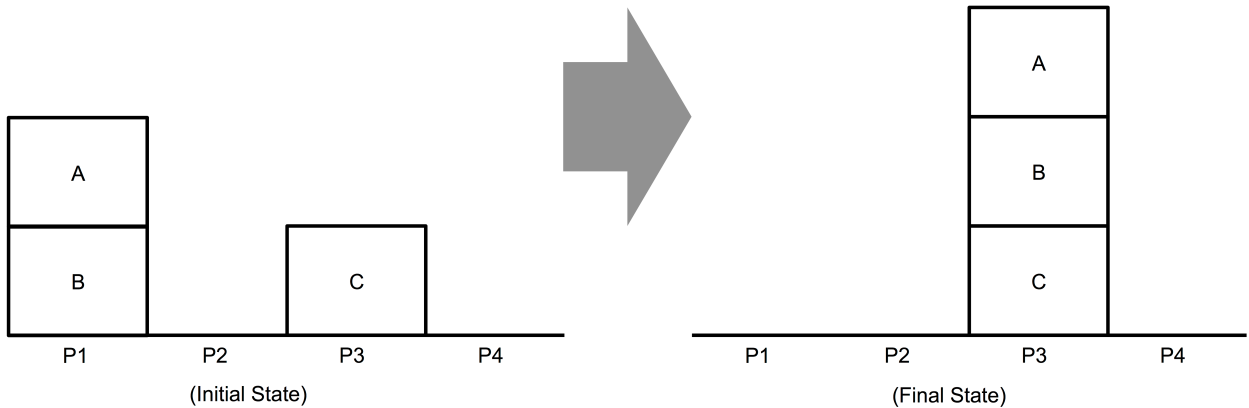

**Fig 3 Example of a blocks world problem**

Two blocks cannot be placed on the same position therefore a block can only be moved to a place if the place is clear. Predicate Clear/1 is true for clear places. The agent can only move a block at a time, and only if the block to be moved does not have any other block on top of it. The predicate On/2 is used to specify the relation between a block and the place in which it is positioned. On(*block*, *place*) means that the specified block is positioned on the specified place.

The states of the world contain only positive instances of the Clear/1 and On/2 predicates, which are accessible to all functional components of the agent through its sensors. The agent's only action schema, Move(*block*, *from*, *to*), moves the specified block, from the specified place (*from*), to the specified place (*to*).

We have made some simplifications in the demonstration to be described. These simplifications will be relaxed in future experiments. First, we assume that the agent's observer component recognizes each time a final state is reached; final states are always successful.

According to the second simplification, each action produces always the same effects. Conditional and imperfect actions were not considered.

Third, we have also assumed that the agent has perfect access to all relevant aspects of the world.

All initial configurations (i.e., 120) were automatically generated and the agent's executor component had to stack the blocks in each of them, which represents 120 instances of the same class of behaviour. The states and the actions of each behaviour were recorded on a file. The 120 files were processed by the defined algorithms, which acquired the defined concepts and generated explanations of the agent's behaviours.

**General results**. Our algorithms have correctly discovered all the described concepts. In particular they discovered the sets of domain entities: propositions, fluent propositions (e.g., On(A, B), Clear(A)), static propositions (e.g., Block(A), Place($P_1$)), and actions (e.g., Move(A, $P_1$, B)).

The agent's observer functional component also discovered the agent's goals:

Goal({On(A, B), On(B, C), Clear(A)})

One might argue that Clear(A) is not a goal; it is a necessity in the sense that it is true in all states of the described world in which the blocks A, B, and C are stacked. Although we have not solved this problem, we have in mind an approach that would generate experiences that would be capable of distinguishing true goals from mere necessities. The experiments would be designed to confirm or disconfirm goals under suspicion. Since the space of all possible experiments would be intractable, the mentioned approach should also provide a way of identifying goals under suspicion. For instance, if a certain proposition is the precondition of any action that achieves one of the agent's goals, maybe that proposition is not a true goal, but a necessity.

Desired propositions are individual goals. Accordingly, the agent's observer component correctly identified the following desired propositions: Desired(On(A, B)), Desired(On(B, C)), Desired(Clear(A)).

The agent's observer component correctly discovered that On(B, C) must be achieved before it can achieve On(A, B): MustPrecede(On(B, C), On(A, B)).

In fact, the agent would never be capable of achieving its goals if this precedence relation is not observed. Moreover, On(B, C) is not a precondition of the actions that achieve On(A, B). The precedence relation arises strictly of the relation between the agent's goals, its capabilities, and possibly the environment.

Given the deduction axiom for the Mandatory/1 relation, the agent's observer component didn't identify any mandatory proposition because, although On(B, C) must precede On(A, B), On(B, C) is also a goal, and goals were not defined as mandatory.

Only to test the deduction axiom and corresponding algorithm, we have artificially added the proposition GoalPrecedingProp before the proposition On(A, B) is achieved. Since GoalPrecedingProp is not a goal, it was identified as a mandatory proposition, which agrees with our preliminary definition:

Mandatory(GoalPrecedingProp).

The agent's observer component was also capable of discovering the generalized action descriptions:

Precond(Move(*block*, *from*, *to*)) = {On(*block*, *from*), Clear(*block*), Clear(*to*)}

PosEffects(Move(*block*, *from*, *to*)) = {On(*block*, *to*), Clear(*from*)}

NegEffects(Move(*block*, *from*, *to*)) = {On(*block*, *from*), Clear(*to*)}

ValidityCondition(Move(*block*, *from*, *to*)) ≡

(Block(*block*) ∧ Place(*from*) ∧ Place(*to*) ∧ *block* ≠ *from* ∧ *block* ≠ *to* ∧ *from* ≠ *to*)

To learn the inequality relationships (≠) between all pairs of variables, it was necessary to equip the agent with sensors for the equality relation (=) and provide it with prior knowledge that, for any two terms, $(X \neq Y) \equiv \neg(X = Y)$.

The following equality relationships among blocks and among table spaces were introduced in all states of all agent behaviours: $A=A$, $B=B$, $C=C$, $P_1=P_1$, $P_2=P_2$, $P_3=P_3$, $P_4=P_4$. These relationships were correctly learned as static relationships.

The algorithm was then allowed to use all static propositions together with the inequality relation and its definition in terms of the equality relation. This allowed the algorithm to correctly use the inequality relationships among all variables, when determining the action validity condition.

**Explaining specific behaviours**. Achieved/3 and Contributed/3 are the relations that provide meaning to specific behaviours. As explained, Achieved(*State*, *Action*, *PropSet*) and Contributed(*State*, *Prop*, *Action*) express the importance of specific actions and specific propositions in specific behaviours. Their importance is always explained in direct or indirect relation to the agent's goals, hence Achieved/3 and Contributed/3, we argue, represent cold *first-person* meanings. They will become true *first-person* meanings when they are consciously felt by the agent, in the phenomenological sense.

The agent's observer component correctly explained all agent behaviours. However, it is impossible and useless to show the explanations of all actions of all 120 observed behaviours. Instead, we describe an example behavioural instance (Fig 4) and we illustrate the explanations discovered by the agent's observer component.

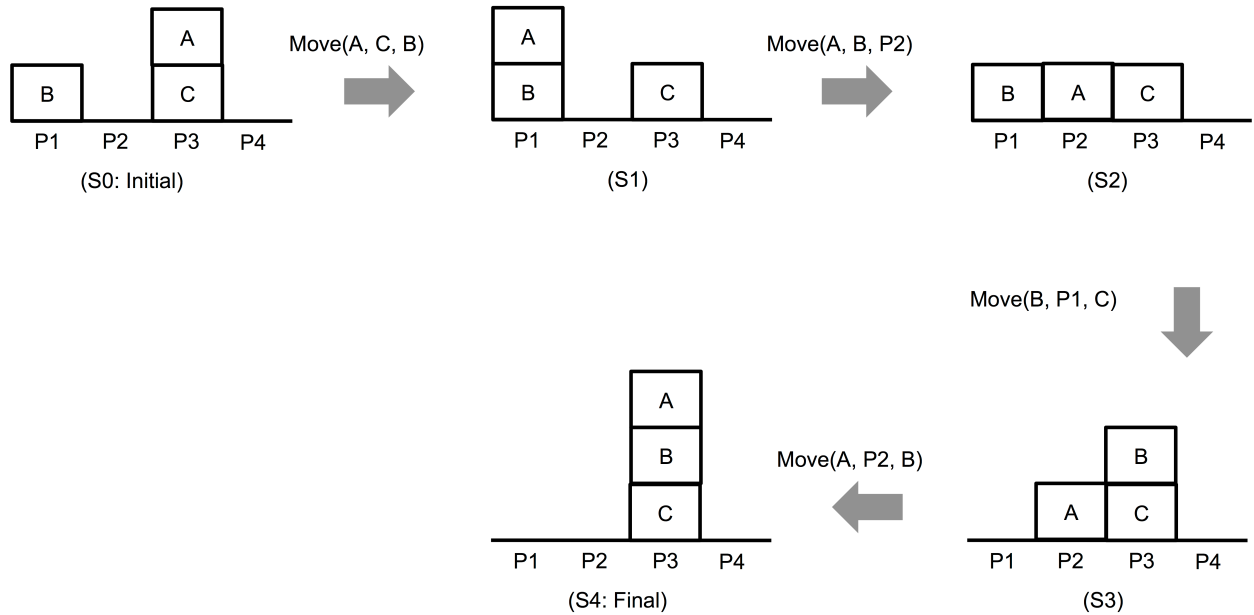

**Fig 4 Example of a behavioural instance**

We start with the *Achieved/3* explanations. The agent was capable of describing the relevant effects of all actions, but we provide only two examples of the generated explanations.

Achieved(S$_1$, Move(A, B, P$_2$), {On(A, P$_2$), Clear(B)})

This means that the action Move(A, B, P$_2$), executed in state S$_1$, led to a state (S$_2$) in which block A is located on place P$_2$, On(A, P$_2$), and block B is clear, Clear(B). These two effects are relevant because the propositions On(A, P$_2$) and Clear(B) were not true in state S$_1$, and because Clear(B) is one of the preconditions of the action Move(B, P$_1$, C), actually executed in state S$_2$, and On(A, P$_2$) is one of the preconditions of action Move(A, P$_2$, B), actually executed in state S$_3$. Although Move(A, P$_2$, B) has not been executed immediately in state S$_2$, the proposition On(A, P$_2$) held true in states S$_2$ and S$_3$, which means it was actually used by the action Move(A, P$_2$, B).

Achieved(S$_2$, Move(B, P$_1$, C), {On(B, C)})

This means that the action Move(B, P$_1$, C), executed in state S$_2$, led to a state (S$_3$) in which the block B is placed on top of block C, On(B, C). This is relevant because, although On(B, C) is not the precondition of any of the actions executed afterwards, it is one of the agent's goals.

The agent's observer component was also capable of finding meaning in the propositions that occur in the states of the agent behaviour. Once again, we provide two examples that illustrate the results.

Contributed(S$_0$, Clear(A), Move(A, C, B))

Clear(A), true in state S$_0$, contributed to the action Move(A, C, B), actually executed in state S$_0$, because it is one of its preconditions.

Contributed(S$_0$, Clear(B), Move(A, C, B))

Clear(B), true in state S$_0$, contributed to the action Move(A, C, B), actually executed in state S$_0$, because it is one of its preconditions.

Achieved/3 and Contributed/3 relationships can be chained to provide an explanation of the whole behaviour.

We have also implemented a small exploratory fraction of the agent's adopter functional component. As can be seen in Fig 4, the agent we have programmed to perform tasks in the blocks world is not an optimal agent. In fact the first two of its actions, in the depicted behaviour, should have been replaced by Move(A, C, $P_2$) or Move(A, C, $P_4$). However, if the description of the agent's actions (Precond/1, PosEffects/1, NegEffects/1 and ValidityCondition/1) and the agent's goals are used in a planning algorithm, the agent will exhibit a better behaviour. We have actually used a planning algorithm to which we have provided the agent's goals and the descriptions of its actions. As expected, the algorithm produced an optimized behaviour. This means that an agent can improve its future performance after it understands the goals that move its behaviour, the description of its capabilities, and the properties of the environment.

## 4   Conclusions

We have argued that it is possible to have computer programs that develop *first-person* meanings of what they do and of what happens to them. Our belief relies on the following more fundamental one: along the evolution of life, there was a stage in which the most sophisticated living beings, although with brains and nervous systems, did not develop *first-person* meanings. As computer programs, they were pure rule-following systems, in the sense that their behaviour was determined by the forces applicable to their inner workings. Yet, in a more recent stage of evolution, those living beings gave rise to other more sophisticated ones, capable of developing *first-person* meanings. That is, a rule-following system with no *first-person* meanings can evolve or develop into a system with *first-person* meanings. We note however that an appropriate test would be a significant advancement.

Damásio [13] suggests that the concerns of an individual arise only if the individual is consciously aware of what he or she does and of what happens to him or her. While fully adopting this position, we extend it by saying that cold awareness is not enough; it is necessary to feel, to be phenomenally aware of what one does and of what happens to us. Varela and Depraz [66], and Colombetti [67] also recognize the primordial importance of the feeling body to consciousness. *First-person* meanings do not arise of some yet to be understood magic trick of pure rule-following; they are created by our rationalization capabilities and by our phenomenal experience.

From the above argument, we propose that for a program to develop *first-person* meanings, it has first to start seeing itself as a goal-directed entity, so that it can explain what happens in terms of its relationship with its goals. Then it must adopt the goals it learns to have, as well as acquired goal-based explanations. For this adoption process, *(i)* the program's future behaviour must reflect the knowledge the program acquires about itself and about its environment and tasks; and *(ii)* the program must feel the importance of objects, events, and situations to its goals, its capabilities, and the constraints and opportunities presented by its environment. The research described in this article contributed to the less demanding problem of the agent becoming aware (in the cold sense) of its goals, its capabilities, its behaviour, and everything that surrounds it, and use that awareness to develop cold *first-person* meanings of objects, events, and situations. We have also made a smaller contribution to enable an agent that has acquired the mentioned knowledge of its goals, of its capabilities, and of the properties of the environment to use that knowledge to shape its future behaviour.

Although we believe that a program may have *first-person* meanings comparable with ours only when it consciously feels, we are firmly convinced that it is possible to start right now with what could be called cold *first-person* meanings.

We proposed a research framework aimed at the development of agents having *first-person* meanings. The proposed framework was shaped by five research assumptions. Finally, we described the research we have recently done and its results.

Tomasello and colleagues [68] describe the way behaviours are understood by a human observer, at three increasing levels of sophistication: the actor behaves autonomously (he/she is the responsible for its behaviour) but no notion of goals is perceived by the observer, the actor is moved by goals, and the actor chooses plans to achieve its goals.

Using our proposal, the agent perceives its own behaviour as goal-directed, the second level of sophistication proposed by Tomasello and colleagues [68] (our agent does not perceive its behaviour as determined by plans, the third level of sophistication).

For these researchers [68], understanding other persons as goal-directed does not happen until a certain stage of development (i.e., 9-month-olds). Although this does not proof our proposal, it is at least interesting that perceiving behaviour as goal-directed is not innate in children, which is consistent with our proposal: the agent takes some time until it discovers that its behaviour may be understood as if it had goals, until it understands its capabilities, until it learns the impact of its environment on the achieving of its goals, and until it develops *first-person* meanings of what it does and of what happens to it.

Another piece of indirect support to the point of view presented in this article comes from Wiggins [7]. According to Wiggins, it seems credible that consciously experienced goals are actually unconsciously generated. Goals assume the most intuitive status that is usually attributed to them only when they enter into consciousness. This possibility seems consistent with our proposal that at first the agent is not aware of having goals. It becomes aware of its goals only when it discovers the goals at the service of which, its behaviour might have happened.

Our research will be pursued in several ways. First of all, the algorithms proposed in this article need to consider all instances of the agent problem solving to be able to learn the agent's goals, the description of its actions and relations between the properties of the environment, which are the basis for the development of *first-person* meanings. We want to design an incremental version of our algorithms following, for example, the work of Baral and Son [69], and Hunter and Delgrande [70, 71] on iterated belief change after sequences of actions and observations.

Then, we intend to drop some simplifying hypotheses we have used to facilitate this work. The agent's observer component should not know *a priori* when the agent's executor component has reached the final state of its task and if the task was successfully accomplished or if anything failed. The agent world should not be fully accessible to its sensors, and both the agent sensors and actuators should not be perfect. Not everything returned by the agent sensors should always exactly match reality; and not all agent actions should have absolutely predictable effects. All aspects of the reported research may be made more robust. We will also expand our work to tackle persistent goals, such as staying alive.

It will also be important that the agent's observer component monitors the changes in the agent's internal state, in addition to just observing what happens in the external world. This might enable the agent to interpret what it does as a strategy to preserve or to achieve desired properties of its internal state.

All the algorithms used in the research presented in this article, although totally domain independent, were especially designed for our problems. It might be insightful to approach all the learning problems (those in which the agent's observer component considers all instances of the same behaviour) using existing general-purpose learning algorithms and compare the results with those of the special-purpose algorithms we have designed.

The vocabulary we have proposed to represent the agent intrinsic understanding of what it does and what happens to it is still not enough. For instance, the agent does not have yet the vocabulary to express astonishment, and it has little capability to express properties of the environment. We will investigate new concepts, to be learned by the agent, which will empower it with capabilities for a richer understanding of what it does and what happens to it.

Another way this work will be pursued will be by investigating better mechanisms by which our computer programs may use the knowledge they acquire about themselves and the environment they inhabit to better shape their future behaviour.

Finally, we will start a research endeavour on what it means for an agent to consciously feel what it does and what surrounds it. A rigorous test must be developed, but its design requires a deeper understanding of what is feeling in the phenomenological sense.

## 5 Acknowledgements

This work was done in the scope of R&D Units 50008 and 50021, partially supported by national funds through Fundação para a Ciência e a Tecnologia (FCT) with contract references UID/EEA/50008/2019 and UID/CEC/50021/2019. We would like to thank Filipe Santos for his review of the logical formalization.